\RequirePackage{fix-cm}
\documentclass[twocolumn]{svjour3}          
\smartqed  
\usepackage{epstopdf}
\usepackage{url}
\usepackage{indentfirst}
\usepackage{graphicx}
\usepackage{amssymb}
\usepackage{times}
\usepackage{subfigure}
\usepackage{amsmath}
\usepackage{multirow}
\usepackage[misc]{ifsym}
\usepackage[numbers]{natbib}
\usepackage{color}
\usepackage{lineno,hyperref}

\usepackage[misc]{ifsym}

\begin{document}

\title{Pseudo-positive regularization for deep person re-identification}

\authorrunning{F. Zhu et al.}        

\author{Fuqing Zhu \and Xiangwei Kong \and Haiyan Fu \and Qi Tian}

\institute{F. Zhu \and X. Kong ({\Letter}) \and H. Fu \\
           School of Information and Communication Engineering, Dalian University of Technology, Dalian 116024, China\\
           e-mail: kongxw@dlut.edu.cn\\
           \at
           F. Zhu\\
           e-mail: fuqingzhu@mail.dlut.edu.cn\\
           \at
           H. Fu\\
           e-mail: fuhy@dlut.edu.cn\\
           \at
           Q. Tian\\
           Department of Computer Science, University of Texas at San Antonio, San Antonio, TX 78249-1604, USA\\
           e-mail: qitian@cs.utsa.edu
}

\date{Received: date / Accepted: date}

\maketitle

\begin{abstract}
  An intrinsic challenge of person re-identification (re-ID) is the annotation difficulty. This typically means 1) few training samples per identity, and 2) thus the lack of diversity among the training samples. Consequently, we face high risk of over-fitting when training the convolutional neural network (CNN), a state-of-the-art method in person re-ID. To reduce the risk of over-fitting, this paper proposes a \textbf{P}seudo \textbf{P}ositive \textbf{R}egularization (PPR) method to enrich the diversity of the training data. Specifically, unlabeled data from an independent pedestrian database is retrieved using the target training data as query. A small proportion of these retrieved samples are randomly selected as the Pseudo Positive samples and added to the target training set for the supervised CNN training. The addition of Pseudo Positive samples is therefore a data augmentation method to reduce the risk of over-fitting during CNN training. We implement our idea in the identification CNN models (\emph{i.e.}, CaffeNet, VGGNet-16 and ResNet-50). On CUHK03 and Market-1501 datasets, experimental results demonstrate that the proposed method consistently improves the baseline and yields competitive performance to the state-of-the-art person re-ID methods.
\end{abstract}

\section{Introduction}
Person re-identification (re-ID) is the task of matching the same person across non-overlapping cameras, which has received increasing research interests in automated surveillance system due to its potential in public security applications. From the perspective of computer vision, the most challenging problem in re-ID is how to correctly match two images (bounding boxes) of the same person under cross scenarios due to the varieties of lighting, pose and viewpoint.

\begin{figure}[t]
  \centering
  \includegraphics[width=0.325\textwidth{},keepaspectratio]{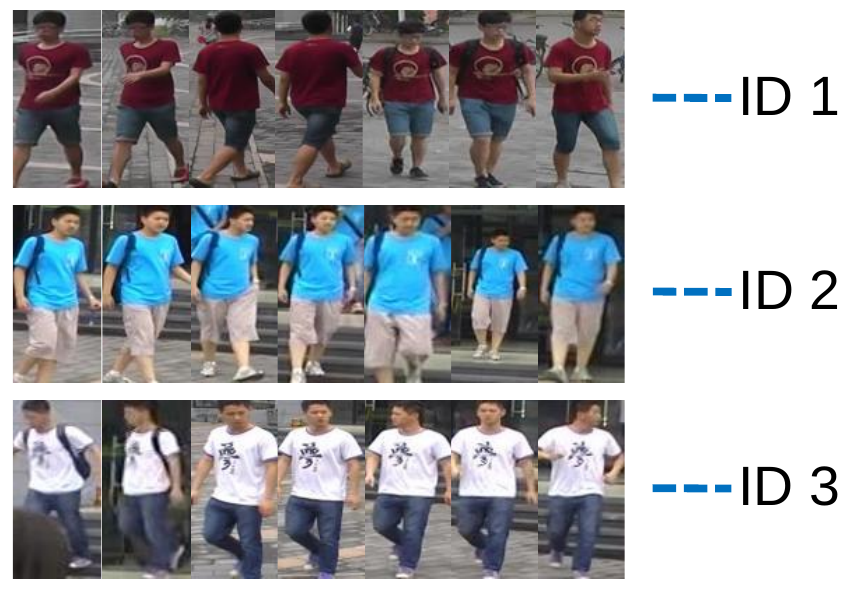}%
  \caption{The training samples of some identities in Market-1501 \cite{zheng2015scalable} dataset. We observe that the intra-class variations are limited.}\label{fig:small_intra}
\end{figure}

Person re-ID lies in between image classification \cite{yang2013feature,yang2015new,yan2016image,chang2016compound} and retrieval \cite{yang2008harmonizing,yang2012multimedia,liu2014sparse,li2014tag,zheng2014packing,zheng2015query,zheng2015tensor,fu2016bhog,zheng2016accurate,zheng2017sift}, which has made a detailed discussion
in \cite{zheng2016person}. Recently, convolutional neural network (CNN) based methods have been record-leading in person re-ID \cite{Xiao2016Learning,zheng2016mars,varior2016gated,varior2016siamese} community. Typically, the underlying requirement of the CNN is a rich amount of samples for each training identity, so that the variation of the intra-class data helps the discriminative learning. However, in person re-ID, a concurrent problem is the difficulty in data annotation: it is not trivial to collect a large amount of cross-camera samples for each training identity. For example, there are 17.2 training samples on average for each identity on the currently largest Market-1501 \cite{zheng2015scalable} dataset, and these existing training samples exhibit very limited intra-class variations as shown in Fig. \ref{fig:small_intra}. When the training data does not have sufficient intra-class variations, there can be high risk of over-fitting. This problem may compromise the discriminative ability of the trained CNN model.

Another motivation of this paper is that there are limited works on how to make use of the unlabeled data for person re-ID. Previous works either focus on unsupervised descriptor \cite{zheng2015scalable,bazzani2014sdalf} design or metric/transfer learning \cite{peng2016unsupervised,kodirov2016person}. In this work, instead, we discuss how to utilize the unlabeled data to improve the discriminative ability of the CNN model. Our work will thus provide possible insight on how to improve the existing CNN model by the low-cost unlabeled data for person re-ID task.

Given the above two considerations, this paper proposes a \textbf{P}seudo \textbf{P}ositive \textbf{R}egularization (PPR) method to reduce the risk of over-fitting during CNN training. The Pseudo Positive samples are defined as pedestrians in very similar appearance but belonging to distinct identities to the positive training samples in each identity. Our method has two characteristics: 1) the Pseudo Positive samples are generated from an independent database\footnote{The independent database is obtained from our collaboration project with Dr. Liang Zheng at the University of Technology Sydney (homepage: http://www.liangzheng.com.cn/). The pedestrian images are captured from several cameras placed in front of a supermarket at Tsinghua University. The database will be made publicly available together with Dr. Liang Zheng's future publication.} (can be regarded as unlabeled data); 2) by adding a small proportion of the Pseudo Positive samples to the fully supervised training data, the training time remains mostly unchanged while more discriminative CNN model can be learned.

More specifically, the Pseudo Positive samples are generated from the independent database using the supervised training samples as queries by nearest neighbor search. The feature representations of both the queries and independent database are the deep features (response of intermediate layer) extracted by the trained baseline CNN model. A small proportion of these retrieved samples are randomly selected as the Pseudo Positive samples and added to the target original training set in each identity for the supervised CNN training, with which the proposed PPR based CNN model can be learned. To summarize, this paper has two contributions.

\begin{itemize}
  \item We propose the \textbf{P}seudo \textbf{P}ositive \textbf{R}egularization (PPR) method to regularize the training process of the CNN model to reduce the risk of over-fitting for person re-ID.
  \item On two large-scale person re-ID datasets (\emph{i.e.}, CUHK03 \cite{li2014deepreid} and Market-1501 \cite{zheng2015scalable}), our method demonstrates consistent improvement over the corresponding fully supervised baseline.
\end{itemize}

The rest of this paper is organized as follows. In Section \ref{sec_2}, we review related work briefly. The proposed PPR method will be described in Section \ref{sec_3}. In Section \ref{sec_4}, extensive results are presented on CUHK03 and Market-1501 datasets. Finally, we conclude this paper in Section \ref{sec_5}.

\section{Related work}\label{sec_2}
This paper focuses on the risk of over-fitting when training the CNN model for person re-ID task. In this section, we will discuss the related work in CNN-based person re-ID methods and some solutions of avoiding over-fitting.

\subsection{CNN-based person re-ID methods}
The CNN-based deep learning model has become popular in computer vision community since Krizhevsky \emph{et al.} \cite{krizhevsky2012imagenet} won ILSVRC2012\footnote{http://image-net.org/challenges/LSVRC/2012/}. The first two re-ID works using deep learning were \cite{yi2014deep} and \cite{li2014deepreid}. There are two types of the CNN models that are commonly employed in person re-ID. The first type is the identification CNN model which is widely used in image classification \cite{krizhevsky2012imagenet} and object detection \cite{girshick2014rich}. The second type is the Siamese CNN model using pedestrian pairs \cite{radenovic2016cnn} or triplets \cite{schroff2015facenet} as input.

The initial CNN-based person re-ID methods employ the Siamese CNN model due to the scale limitation of the person re-ID datasets, \emph{e.g.}, VIPeR \cite{gray2008viewpoint}, iLIDS \cite{zheng2009associating}, PRID 2011 \cite{hirzer2011person}, CUHK01 \cite{li2012human}, CUHK02 \cite{li2013locally}, OPeRID \cite{liao2014open}, RAiD \cite{das2014consistent} and PRID 450S \cite{roth2014mahalanobis}. Yi \emph{et al}. \cite{yi2014deep} employ a Siamese CNN model, in which the pedestrian image is partitioned into three overlapping horizontal parts, and then the horizontal parts go through two convolutional layers and finally are fused for this pedestrian image. One patch matching layer and one maxout-grouping layer are added in the Siamese CNN model by Li \emph{et al.} \cite{li2014deepreid}. The patch matching layer is used to learn the displacement of horizontal stripes in across-view images, while the maxout-grouping layer is used to boost the robustness of patch matching. Ahmed \emph{et al.} \cite{ahmed2015improved} improve the Siamese CNN model to learn the cross-image representation via computing the neighborhood distance. Wu \emph{et al.} \cite{wu2016personnet} design a network called PersonNet, which deepen the neural networks using convolutional filters of smaller sizes. Varior \emph{et al.} \cite{varior2016gated} propose capturing effective subtle patterns by inserting a gating function after each convolutional layer. Varior \emph{et al.} \cite{varior2016siamese} merge the long short-term memory (LSTM) modules into a Siamese model. The LSTMs process image parts sequentially, and the spatial connections can be learned to increase the discriminative ability of the CNN model. Liu \emph{et al.} \cite{liu2016end} propose focusing on the important local parts of an input image pair adaptively by integrating a soft attention based model. Cheng \emph{et al.} \cite{cheng2016person} design a triplet based Siamese CNN model which takes three pedestrian images as input. For each image, four overlapping body parts are partitioned after the first convolutional layer and fused with a global representation in the fully connected layer. Su \emph{et al.} \cite{Su2016Deep} propose an attribute prediction using an independent dataset and an attributes triplet loss trained on labeled datasets for person re-ID.

Another potentially effective strategy is the identification CNN model, which makes full use of the pedestrian identity annotations. The identification CNN model is more suitable for practical application in large-scale person re-ID. The Siamese CNN model considers pairwise or triplet labels which is a weak label in person re-ID, that is telling whether the image pair belongs to the same identity or not. Xiao \emph{et al.} \cite{Xiao2016Learning} train identities from multiple datasets with a SoftMax loss in the identification CNN model. The impact score is proposed for each fully connected neuron, and a domain guided dropout is imposed based on the impact score. The learned deep features yield an excellent re-ID accuracy. Zheng \emph{et al.} \cite{zheng2017unlabeled} mix the unlabeled samples generated by generative adversarial network (GAN) \cite{radford2015unsupervised} with the original labeled real training images for semi-supervised learning to improve the discriminative ability of the identification CNN model for person re-ID. Lin \emph{et al.} \cite{lin2017improving} propose an attribute person recognition (APR) CNN model, which learns the pedestrian identity embedding and predicts the pedestrian attributes, simultaneously. Sun \emph{et al.} \cite{sun2017svdnet} propose to decorrelate the learned weight vectors using singular vector decomposition (SVD) to improve the discriminative learning of the identification CNN model for person re-ID. On larger person re-ID datasets, such as MARS \cite{zheng2016mars} and PRW \cite{zheng2017person}, the identification CNN model achieves excellent accuracy without any external training sample selection process. In the recent person re-ID survey \cite{zheng2016person}, some baseline results are presented for both the Siamese and identification CNN models on Market-1501 \cite{zheng2015scalable} dataset, from which the identification CNN model is superior to Siamese CNN model.

\subsection{Solutions of avoiding over-fitting}
Avoiding over-fitting is a major challenge against training discriminative CNN models. Some early solutions include reducing the complexity of network by reducing or sharing parameters \cite{lecun1998gradient}, stopping training process early \cite{plaut1986experiments}, \emph{etc}. Inspired by regularization method in \cite{krizhevsky2009learning}, various regularization methods have been widely adopted in the existing deep neural network recently, such as Data Augmentation \cite{krizhevsky2012imagenet}, Dropout \cite{hinton2012improving}, DropConnect \cite{wan2013regularization} and Stochastic Pooling \cite{zeiler2013stochastic}. The main idea of Data Augmentation is to generate more training data, such as the operation of randomly cropping, rotating and flipping the input images. Dropout discards a part of neuron response randomly and updates the remaining weights in each mini-batch iteration. DropConnect only updates a randomly selected subset of weights. Stochastic Pooling randomly samples one input as the pooling result in probability in the training process of the network. Besides, some other regularization works focus on noisy labeling \cite{vincent2010stacked, van2013learning, van2014marginalizing}, which can be regarded as a special Data Augmentation method. In \cite{sukhbaatar2014training,xie2016disturblabel}, noisy data produced by randomly disturbing the labels of training data is added to the original training set for CNN training in the context of generic image classification.
\begin{figure}[t]
  \centering
  \includegraphics[width=0.475\textwidth{},keepaspectratio]{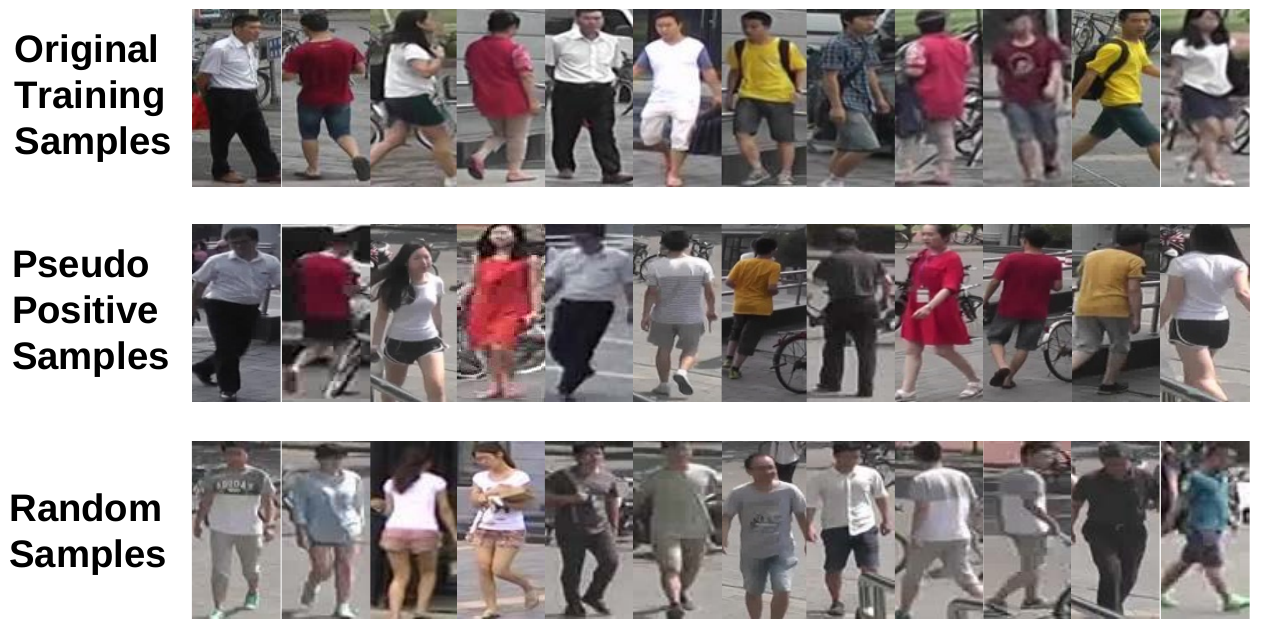}%
  \caption{Examples of the original training, Pseudo Positive and random samples. For each column, the ID of the pedestrian samples is same during CNN training.}\label{fig:sample_pos}
\end{figure}

In this paper, the proposed method departs from \cite{sukhbaatar2014training,xie2016disturblabel} in two aspects. First, this work focuses on person re-ID, a task in which the number of training samples for each identity is much fewer than that in generic image classification. So randomly disturbing the training labels, even a small proportion, will in effect deteriorate the training process (to be evaluated in our experiment). Second, instead of label disturbance, in our work some unlabeled samples retrieved from an independent database are used for Pseudo Positive training samples. This method achieves a delicate design to enrich data diversity while avoiding too much data pollution. Some examples of both the Pseudo Positive and randomly selected (in spirit consistent with \cite{sukhbaatar2014training,xie2016disturblabel}) are listed in Fig. \ref{fig:sample_pos}.

\begin{figure*}[t]
  \centering
  \includegraphics[width=0.75\textwidth{},keepaspectratio]{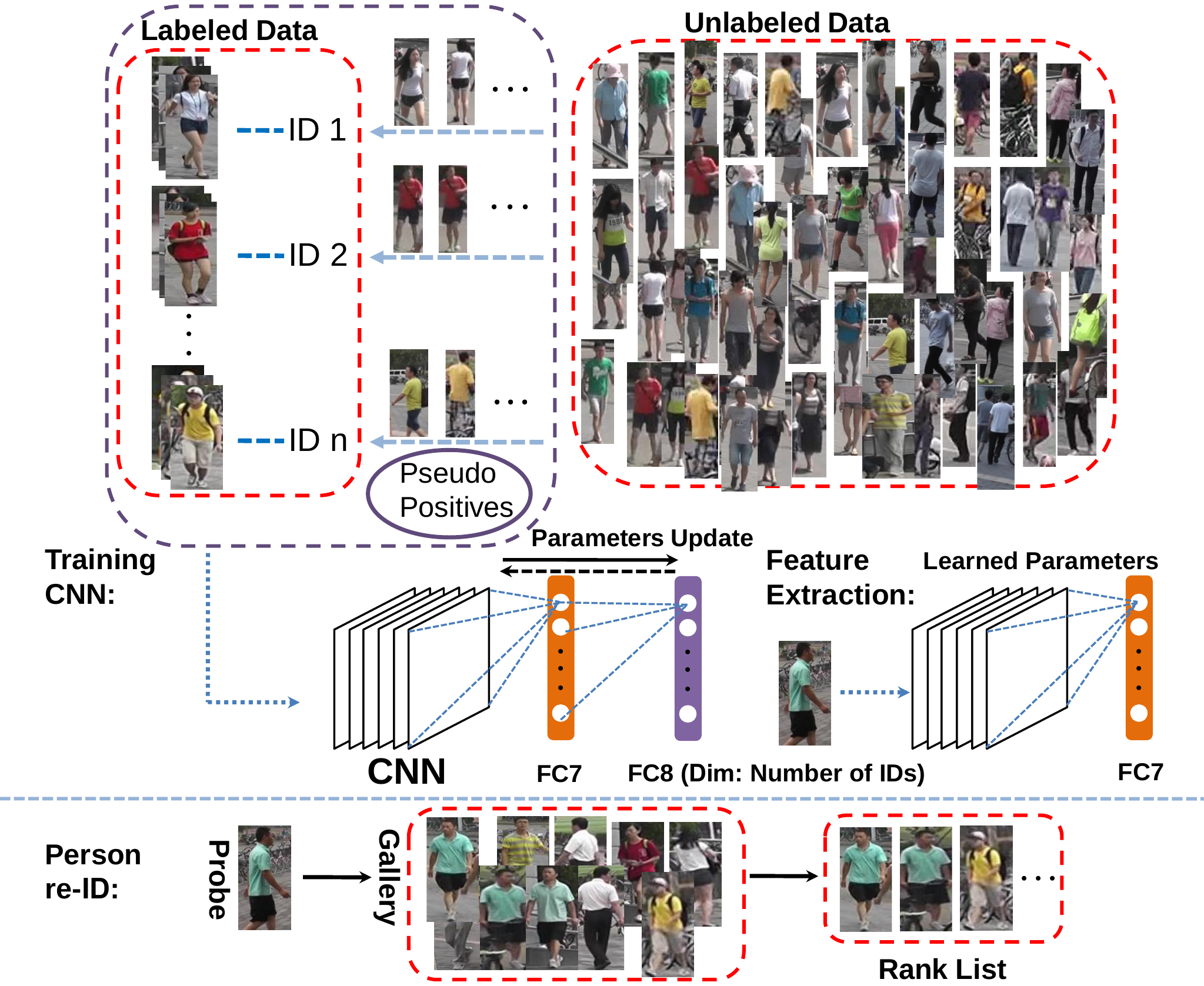}%
  \caption{The framework of the proposed method based on CNN model for person re-ID.}\label{fig:base_im}
\end{figure*}

\section{The proposed method}\label{sec_3}
In this section, we elaborate the proposed Pseudo Positive Regularization (PPR) method based on the identification CNN model for person re-ID. We first describe the baseline. Then, we show the proposed method including the generation of Pseudo Positive samples and the training process of the network. Fig. \ref{fig:base_im} illustrates the framework\footnote{\textbf{Note:} we just take the CaffeNet \cite{krizhevsky2012imagenet} as an example in Fig. \ref{fig:base_im}.} of the proposed PPR method based on CNN model for person re-ID.

\subsection{Baseline}\label{sec:base_l}
According to the results in the recent person re-ID survey \cite{zheng2016person}, the identification CNN model \cite{krizhevsky2012imagenet,he2016deep} outperforms the verification CNN model \cite{ahmed2015improved,li2014deepreid} significantly on Market-1501 \cite{zheng2015scalable} dataset. Because the former model makes full use of the pedestrian identity labels, while the latter model only uses weak labels. So we adopt the identification CNN model as the baseline in this paper. The baseline is an ID-discriminative Embedding using CNN architecture to train the re-ID model in identification mode. Specifically, we use the backbone CNN models (\emph{i.e.}, CaffeNet \cite{krizhevsky2012imagenet}, VGGNet-16 \cite{simonyan2014very} and ResNet-50 \cite{he2016deep}), except editing the last fully connected layer to have the same number of neurons as the number of distinct identities in the training set. We refer readers to the related papers for the detailed network descriptions of CNN models. The identification CNN model is fine-tuned from the ImageNet \cite{deng2009imagenet} pre-trained model for learning a embedding in the pedestrian subspace to discriminate different identities. A discriminative embedding in the person subspace is learned to discriminate different identities during CNN training. After that, the trained CNN model is used as a feature extractor for the pedestrian images in probe (query) and gallery (database) sets.

\textbf{During training,} the training set is denoted as ${\mathcal{D}} = \left\{ {{{\bf{x}}_i},{{y}_i}} \right\}_{i = 1}^N$, in which the pedestrian image is ${{{\bf{x}}_i}}$, and the ID is $y_i$.  Batches of the training samples are fed into the data layer as input of the CNN model. Each sample is composed of a pedestrian image and its associated ID. The training pedestrian images are first resized to 256$\times$256, then randomly cropped into a fixed size\footnote{The size is 227$\times$227 for CaffeNet \cite{krizhevsky2012imagenet}, while the size is 224$\times$224 for VGGNet-16 \cite{simonyan2014very} and ResNet-50 \cite{he2016deep}.} with random horizontal flipping and shuffled before fed into the data layer of CNN model for each iteration. The mean image is subtracted from all training images. The goal is to train an identification CNN model $\mathbb{M}$, which can be regarded as a mapping ${\mathop{\rm f}\nolimits} \left( {{\mathbf{x}},{\mathbf{\theta }}} \right)\in \mathbb{R}^{C}$, where $\mathbf{\theta }$ represents the parameters of each layer in the network. The $\mathbf{\theta }$ is updated using Stochastic Gradient Descent (SGD) \cite{bottou2010large} algorithm in each mini-batch iteration. The $t$-th iteration updates the current parameters $\mathbf{\theta }_t$ as following formula:
\begin{equation}
{{\bf{\theta }}_{t + 1}} = {{\bf{\theta }}_t} + \gamma  \cdot \frac{1}{{\left| {{{\mathcal{D}}_t}} \right|}}\sum\limits_{\left( {{\bf{x}},y \in {{\mathcal{D}}_t}} \right)} {{\nabla _{{{\bf{\theta }}_t}}}\left[ {l\left( {{\bf{x}},y} \right)} \right]}
\end{equation}
where $\gamma$ is learning rate, ${{\mathcal{D}}_t}$ is a mini-batch randomly selected from the training set ${\mathcal{D}}$, $\nabla$ is gradient operation and $l$ is the SoftMax loss function. The SoftMax Loss guides the learning process of the CNN model while the convergence of network must be guaranteed.

\textbf{During testing}, the trained CNN model $\mathbb{M}$ is utilized as a feature extractor. Given an input image, we extract the fully connected or pooling layer activations\footnote{\textbf{Note:} CaffeNet \cite{krizhevsky2012imagenet} and VGGNet-16 \cite{simonyan2014very} are the penultimate fully connected layer, while ResNet-50 \cite{he2016deep} is the last pooling layer.}. The similarity between probe (query) and gallery (database) images is evaluated as the Euclidean distance of the corresponding deep features. By sorting the distance, the final person re-ID result could be obtained based on the rank list.

\subsection{Pseudo Positive Regularization}\label{sec:improv}
PPR utilizes the unlabeled data to reduce the risk of over-fitting during the training process of the CNN model. It provides an effective strategy to improve the discriminative ability of the existing CNN model by the low-cost unlabeled data. The motivation of Pseudo Positive samples generation from unlabeled data is to enrich the diversity of the training samples for each identity.

The Pseudo Positive samples are retrieved from unlabeled independent database ${{\mathcal{D}}_c}$ using the original supervised training samples ${\mathcal{D}}$ as queries. The feature of both queries and unlabeled samples are extracted by the trained baseline CNN model $\mathbb{M}$ described in Section \ref{sec:base_l}. For each labeled sample ${{\bf{x}}_i}$ in training set ${\mathcal{D}}$, we search a nearest neighbor sample ${{\bf{x}}_i}'$ from the unlabeled independent database ${{\mathcal{D}}_c}$ by (\ref{Eq:dist}) based on Euclidean distance,
\begin{equation}
\mathop {\min }\limits_{{{\bf{x}}_i}'} {\left\| {{\mathop{\rm F}\nolimits} \left( {{{\bf{x}}_i}} \right) - {\mathop{\rm F}\nolimits} \left( {{{\bf{x}}_i}'} \right)} \right\|_2},
\label{Eq:dist}
\end{equation}
where ${\mathop{\rm F}\nolimits} \left(  \cdot  \right)$ is the feature representation of the sample.

The proposed method aims to enrich the diversity of training data while avoiding too much data pollution. In the experiment, we only adopt a proportion of these retrieved samples as Pseudo Positive samples for each identity. The generated Pseudo Positive set is denoted as ${{\mathcal{D}}_p} = \left\{ {{{\bf{x}}_i}',{{{y}}_i}'} \right\}_{i = 1}^M$. The ID ${{{{y}}_i}'}$ of Pseudo Positive sample ${{{\bf{x}}_i}'}$ is same as ID ${{{{y}}_i}}$ of training sample ${{{\bf{x}}_i}}$.

\textbf{During training,} the samples of both Pseudo Positive set ${{\mathcal{D}}_p}$ and original training set ${\mathcal{D}}$ are fed into the data layer for training new CNN model $\mathbb{M}'$. The parameters setting of each layer and the training process of the CNN model in PPR is same as the baseline for fair comparison.

\textbf{During testing,} the trained new CNN model $\mathbb{M}'$ is utilized as a feature extractor. The deep feature is used for performing person retrieval in the gallery set as same as the baseline.

Let us discuss the possible insights on how to improve the existing identification CNN model based on the proposed PPR method for person re-ID. Considering the formulation of the identification CNN model, given a set of training samples, the objective would be to learn an optimal mapping which maps each input pedestrian sample to its corresponding ground-truth ID. The richer training samples, the more excellent generalization ability of the trained CNN model can be achieved. For each identity, the Pseudo Positive and original training samples have high similarity in feature representation, so the combination of the two types of samples covers a larger proportion of natural distribution. Moreover, PPR has a larger appearance diversity compared with the traditional Data Augmentation (\emph{e.g.}, horizontally flipping, random crops and color jittering) methods. The Pseudo Positive samples are crawled from unlabeled independent database, while the traditional Data Augmentation is operated on the original images. It is that there is no new samples addition essentially in the traditional Data Augmentation method. The new generated Pseudo Positive samples can reduce the impacts of training data bias by introducing these extra types of variability during the training process of the CNN model. PPR results in a similar effect of increasing training data coverage at every hidden layer. Obviously, this special Data Augmentation way of the proposed method reduces the risk of over-fitting during CNN training which makes it possible to learn a more excellent generalization mapping.

\section{Experiments}\label{sec_4}
In this section, we first describe the datasets and evaluation protocol. Then, we show the experimental results to demonstrate the effectiveness of the proposed method.

\subsection{Datasets and evaluation protocol}
This paper evaluates the performance of the baseline and PPR on the currently largest person re-ID datasets: CUHK03 \cite{li2014deepreid} and Market-1501 \cite{zheng2015scalable}, which are close towards realistic situations. The CUHK03 dataset contains 13,164 bounding boxes of 1,360 identities collected from six surveillance cameras, in which each identity is observed by two disjoint camera views and has an average of 4.8 bounding boxes in each camera view. There are two bounding box generation versions, which are manually labeled and automatically detected by the pedestrian detector DPM \cite{felzenszwalb2010object}, respectively. We evaluate the automatically ``detected'' version in our experiment. Following the protocol in \cite{li2014deepreid}, 1,360 identities are split into 1,160 identities for training, 100 identities for validation and 100 identities for testing. We report the averaged result after training/testing 20 times and use the single shot setting on CUHK03 dataset. The Market-1501 dataset contains 32,668 bounding boxes of 1,501 identities. The generation of bounding boxes is automatically detected by the pedestrian detector DPM \cite{felzenszwalb2010object} completely. Following the protocol in \cite{zheng2015scalable}, 1,501 identities are split into 751 identities for training and 750 identities for testing. We use 90\% of the training data for training CNN model and the rest of 10\% for validation on Market-1501 dataset. The testing process is performed in the cross-camera mode. We choose these two re-ID datasets due to their scales, for which scalable retrieval methods are of great needs.

\setlength{\tabcolsep}{2pt}
\begin{table}[t]
  \centering
  \caption{The rank-1 accuracy (\%) and mAP (\%) of the baseline with different CNN models on CUHK03 and Market-1501 datasets, respectively.}
  \begin{tabular}{l|cc|cc} \hline
  \multirow{2}{*}{Model} & \multicolumn{2}{c}{CUHK03} & \multicolumn{2}{|c}{Market-1501}\\ \cline{2-5}
  & rank-1 & mAP & rank-1 & mAP\\ \hline
  CaffeNet \cite{krizhevsky2012imagenet} & 53.90 & 60.11 & 56.03 & 32.41\\
  VGGNet-16 \cite{simonyan2014very} & 52.60 & 58.94 & 65.38 & 40.86\\
  ResNet-50 \cite{he2016deep} & 54.50 & 60.72 & 72.54 & 46.00\\ \hline
  \end{tabular}
  \label{table:base_line}
\end{table}

In the experiments, we adopt the Cumulated Matching Characteristics (CMC) curve\footnote{The rank-1 accuracy is shown when the CMC curve is absent.} and mean Average Precision (mAP) for evaluation on CUHK03 and Market-1501 datasets. The CMC curve shows the probability that a query identity appears in the rank list of different sizes. This evaluation protocol is generally believed to focus on precision. In case of there is only one ground-truth match for a given query, the precision and recall are the same issue. However, if multiple ground-truths exist, the CMC is biased because recall is not considered. For the CUHK03 and Market-1501 datasets, there are several cross-camera ground-truths for each query. Therefore, the mAP is more suitable to evaluate the overall person re-ID performance. It considers both the precision and recall in the rank list, thus providing a more comprehensive evaluation.

The CAFFE \cite{Jia2014Caffe} package is adopted to evaluate the baseline and the proposed method with different CNN models: CaffeNet \cite{krizhevsky2012imagenet}, VGGNet-16 \cite{simonyan2014very} and ResNet-50 \cite{he2016deep}, respectively. We use stochastic gradient descent with momentum 0.9. The weight decay is 0.0005. On CUHK03, the initial learning rate is set to 0.001 and reduced by a factor of 0.1 after each 5 epochs. Training is done after 25 epochs. On Market-1501, the initial learning rate is set to 0.001 and reduced by a factor of 0.1 after each 15 epochs. Training is done after 75 epochs. The division of training and testing set on CUHK03 and Market-1501 is followed by \cite{li2014deepreid} and \cite{zheng2015scalable}.

\subsection{Experimental results}
In the following subsections, we first evaluate the baseline (described in Section \ref{sec:base_l}). Then, we report the results of the proposed method (described in Section \ref{sec:improv}) and analyze the experimental results. Finally, we show the results of comparison with some state-of-the-art person re-ID methods on CUHK03 and Market-1501 datasets.

\subsubsection{Evaluation of the baseline}
Experimental results of the baseline on the two re-ID datasets are shown in Table \ref{table:base_line}. We observe that a very competitive re-ID accuracy can be achieved by the baseline. Note that the baseline exceeds many previous works \cite{liao2015person,varior2016siamese,zhang2016learning}. Specifically, on Market-1501 dataset, we achieve rank-1 accuracy of 72.54\% using ResNet-50 model. (Note: More previous re-ID performance indicators on Market-1501 dataset have been summarized in the survey \cite{zheng2016person}.)

\setlength{\tabcolsep}{2.0pt}
\begin{table}[t]
  \centering
  \caption{The rank-1 accuracy (\%) and mAP (\%) comparisons of the baseline and PPR for various numbers of Pseudo Positive samples with different CNN models on CUHK03 dataset. $K(\cdot)$ denotes the number of Pseudo Positive samples.}
  \begin{tabular}{l|cc|cc|cc} \hline
  \multirow{2}{*}{Methods} & \multicolumn{2}{c|}{CaffeNet \cite{krizhevsky2012imagenet}} & \multicolumn{2}{c|}{VGGNet-16 \cite{simonyan2014very}} & \multicolumn{2}{c}{ResNet-50 \cite{he2016deep}}\\\cline{2-7}
  & rank-1& mAP& rank-1& mAP& rank-1& mAP\\ \hline
  Baseline & 53.90 & 60.11& 52.60& 58.94 &54.50& 60.72 \\ \hline
  PPR [$K$(600)] & 54.00 & 60.39& \textbf{54.75} & \textbf{60.87} & 55.45  & 61.52\\
  PPR [$K$(700)] & \textbf{54.55} & \textbf{60.68}& 53.95 & 60.48& 55.30  & 61.42\\
  PPR [$K$(800)] & 54.10 & 60.43& 52.90 & 59.35& \textbf{55.75}  & \textbf{61.76}\\
  PPR [$K$(1,200)] & 52.55 & 58.76& 51.15 & 57.23 & 53.70 & 59.96\\ \hline
  \end{tabular}\label{table:improv_1}
\end{table}

\setlength{\tabcolsep}{2.0pt}
\begin{table}[t]
  \centering
  \caption{The rank-1 accuracy (\%) and mAP (\%) comparisons of the baseline and PPR for various numbers of Pseudo Positive samples with different CNN models on Market-1501 dataset. $K(\cdot)$ denotes the number of Pseudo Positive samples.}
  \begin{tabular}{l|cc|cc|cc} \hline
  \multirow{2}{*}{Methods} & \multicolumn{2}{c|}{CaffeNet \cite{krizhevsky2012imagenet}} & \multicolumn{2}{c|}{VGGNet-16 \cite{simonyan2014very}} & \multicolumn{2}{c}{ResNet-50 \cite{he2016deep}}\\\cline{2-7}
  & rank-1& mAP& rank-1& mAP& rank-1& mAP\\ \hline
  Baseline & 56.03 & 32.41& 65.38 & 40.86 & 72.54 & 46.00 \\ \hline
  PPR [$K$(500)] & \textbf{57.16} & 32.60& \textbf{66.03} & \textbf{41.75}& 73.31 & 47.36 \\ 
  PPR [$K$(1,000)] & 56.35 & \textbf{33.11} & 65.77 & 41.05& \textbf{73.87} & \textbf{47.79} \\ 
  PPR [$K$(2,000)] & 55.11 & 31.16 & 62.77 & 38.98 & 72.51 & 45.89 \\ \hline
  \end{tabular}\label{table:improv_2}
\end{table}

\subsubsection{Evaluation of the proposed method}
The experimental setup of PPR is same as the baseline for fair comparison.

\makeatother
\begin{figure*}[t]
  \centering
  \subfigure[CaffeNet \cite{krizhevsky2012imagenet}]{\label{fig:cmc_cuhk1}%
  \includegraphics[width=0.32\textwidth{}]{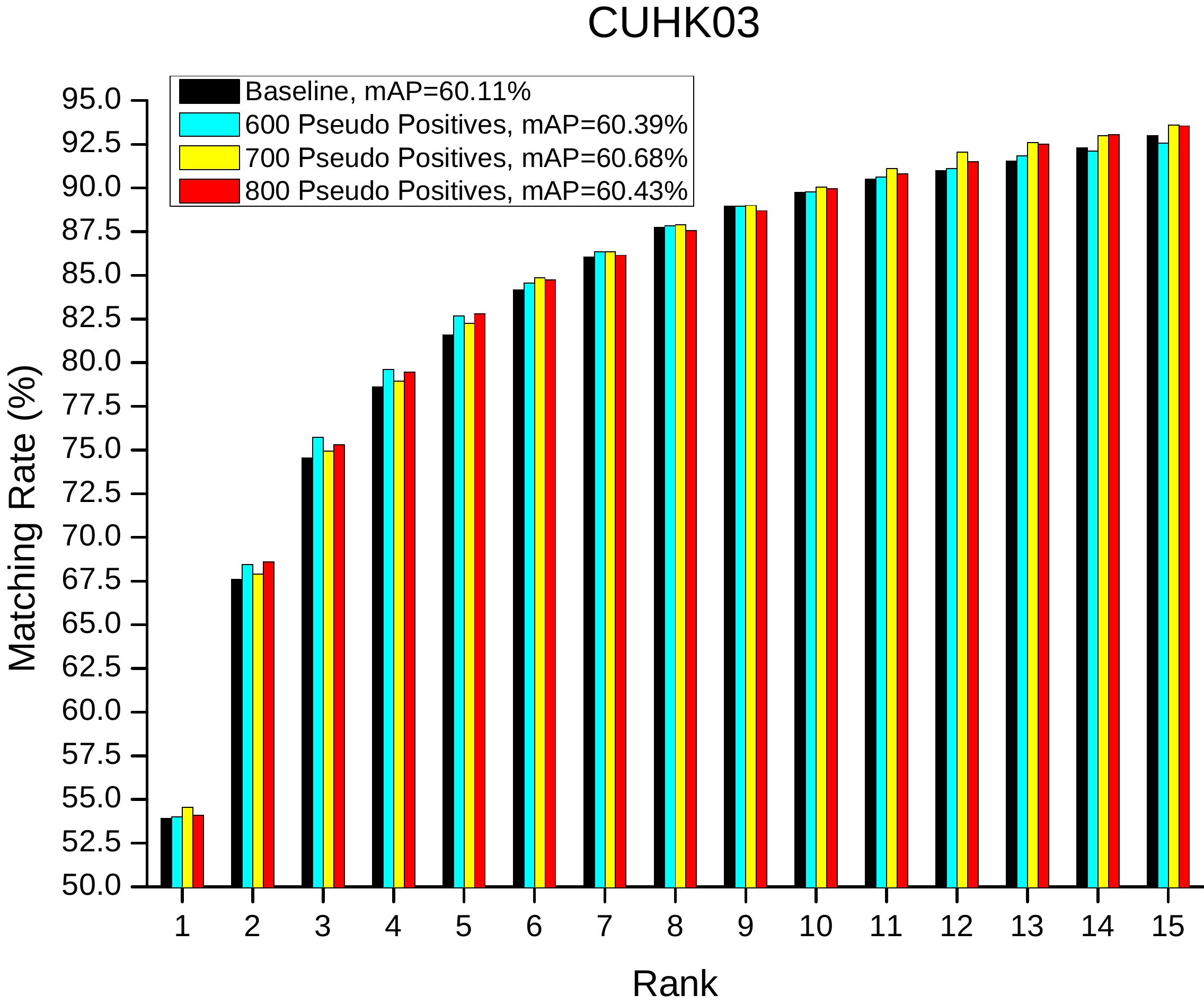}}
  \hspace{0.01in}
  \subfigure[VGGNet-16 \cite{simonyan2014very}]{\label{fig:cmc_cuhk2}%
  \includegraphics[width=0.32\textwidth{}]{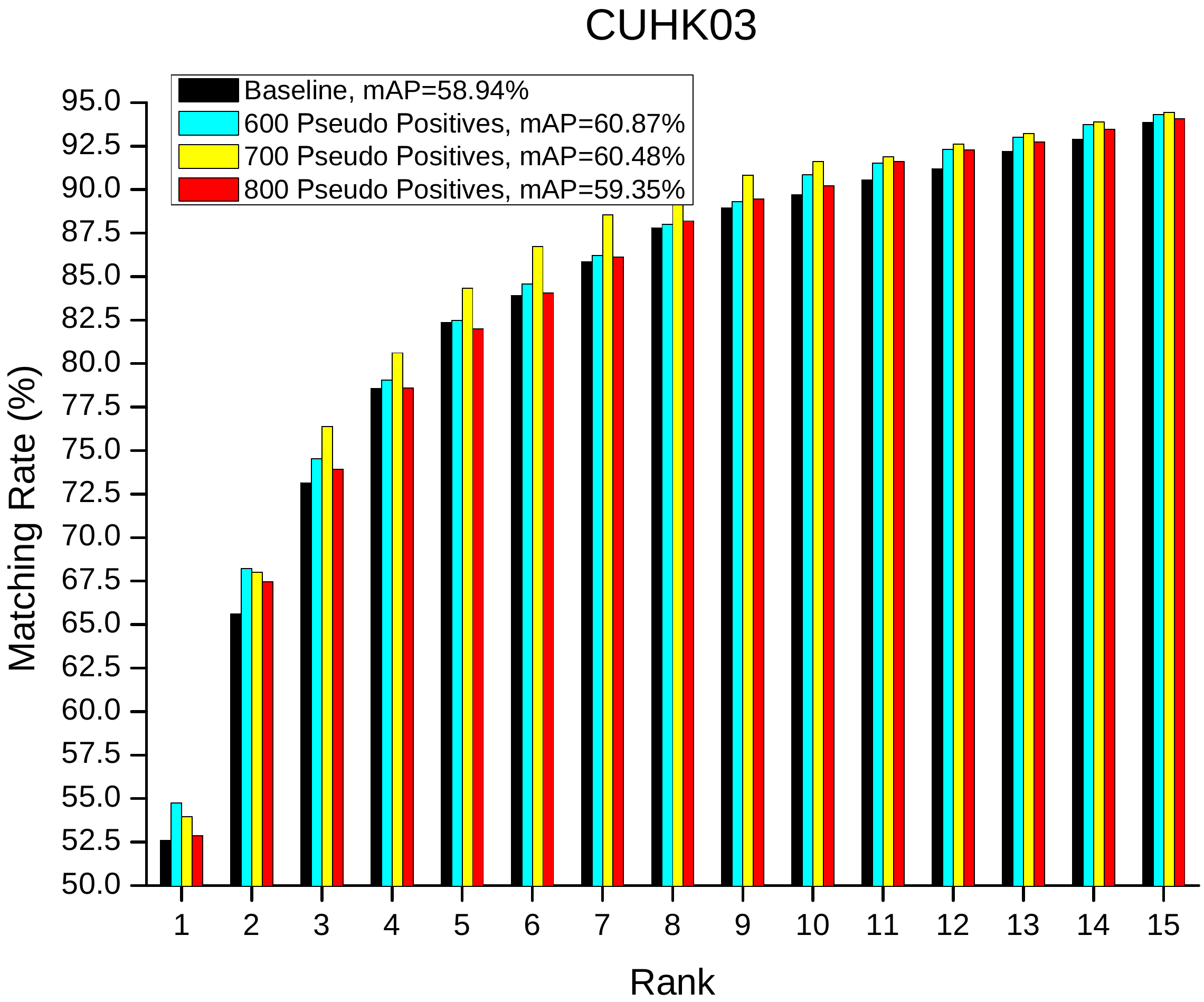}}
  \hspace{0.01in}
  \subfigure[ResNet-50 \cite{he2016deep}]{\label{fig:cmc_cuhk3}%
  \includegraphics[width=0.32\textwidth{}]{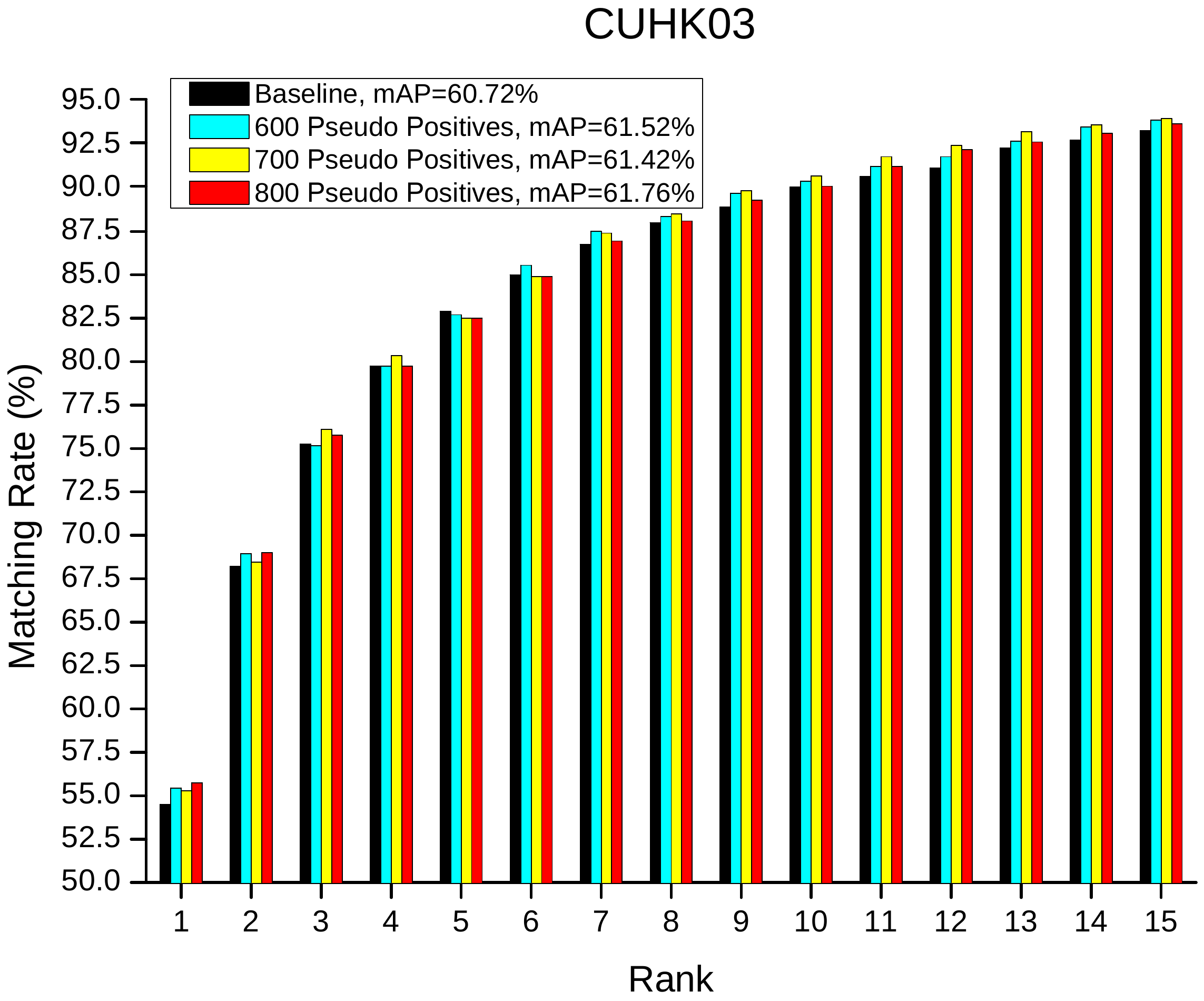}}
  \caption{The CMC comparison of the baseline and PPR on CUHK03 dataset.}
  \label{fig:cmc_cuhk}
\end{figure*}

\makeatother
\begin{figure*} [t]
  \centering
  \subfigure[CaffeNet \cite{krizhevsky2012imagenet}]{\label{fig:cmc_market1}%
  \includegraphics[width=0.32\textwidth{}]{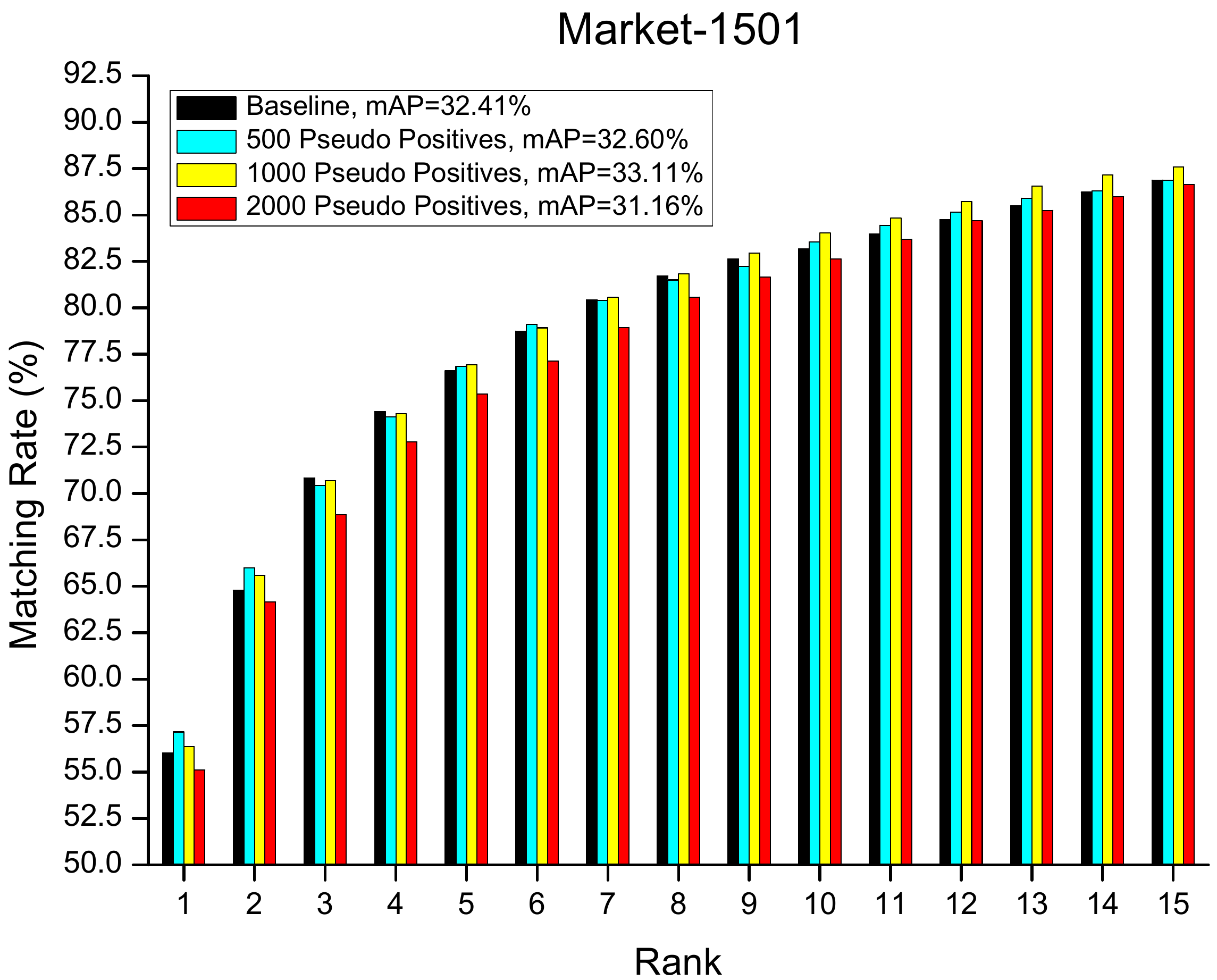}}
  \hspace{0.01in}
  \subfigure[VGGNet-16 \cite{simonyan2014very}]{\label{fig:cmc_market2}%
  \includegraphics[width=0.32\textwidth{}]{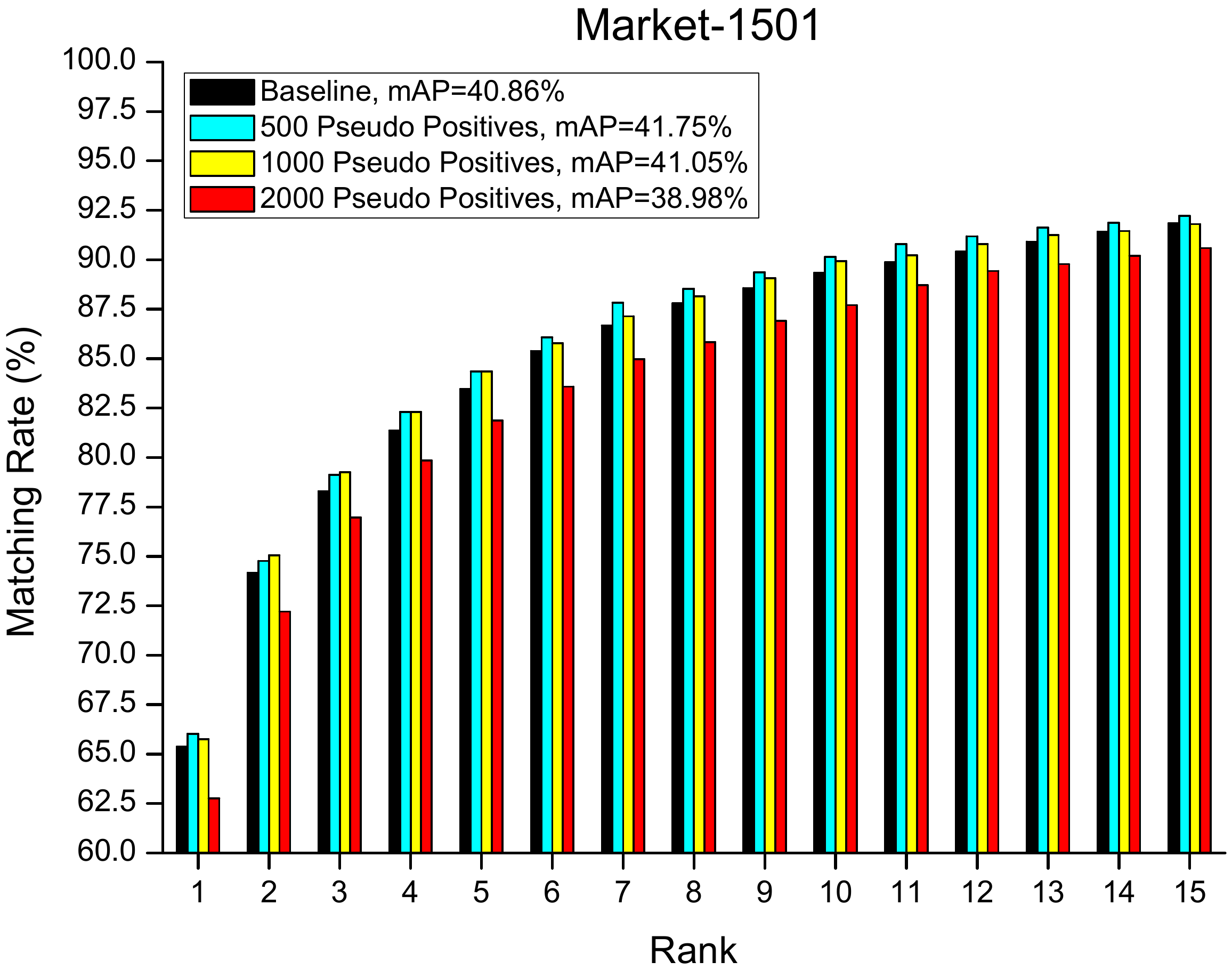}}
  \hspace{0.01in}
  \subfigure[ResNet-50 \cite{he2016deep}]{\label{fig:cmc_market3}%
  \includegraphics[width=0.32\textwidth{}]{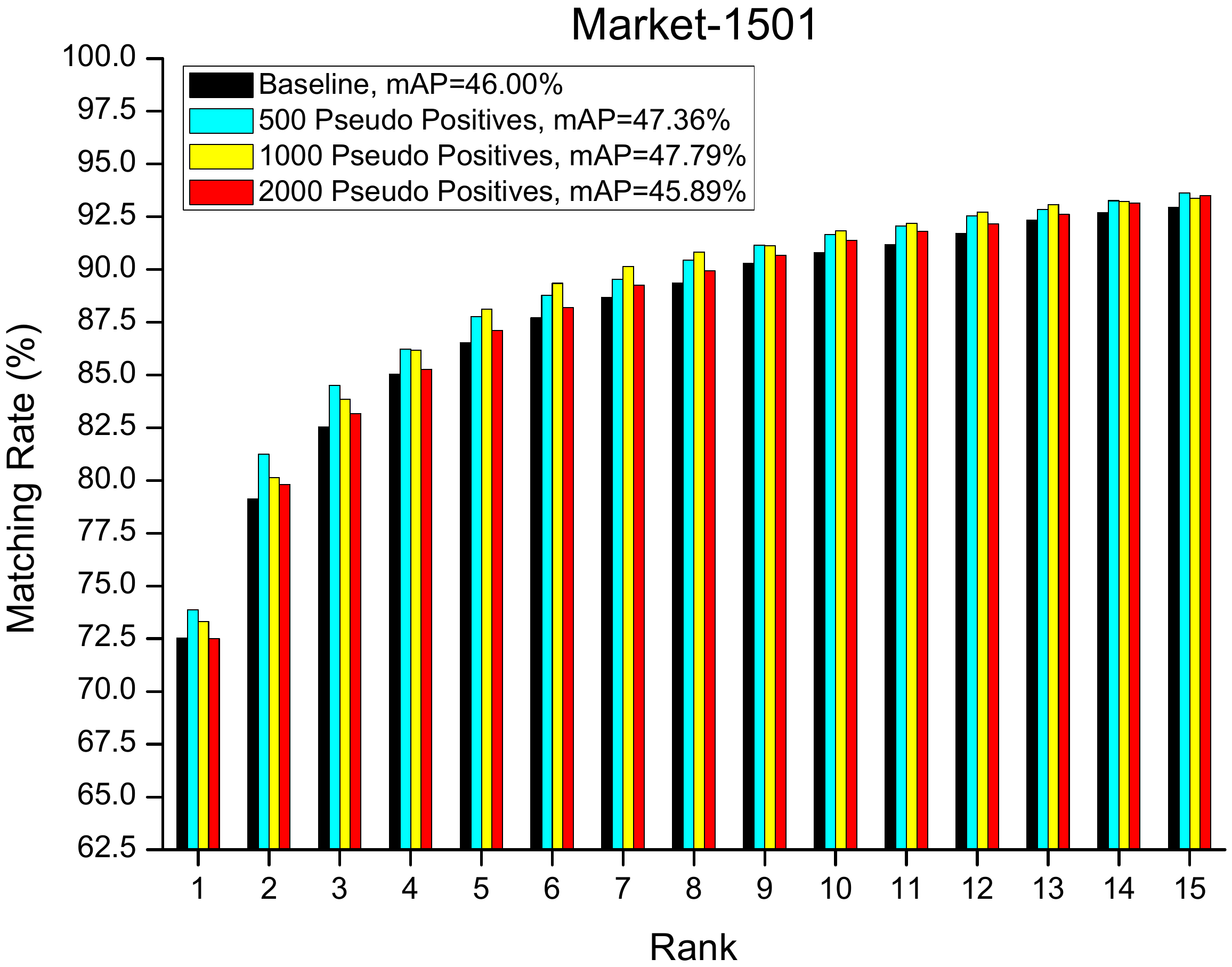}}
  \caption{The CMC comparison of the baseline and PPR on Market-1501 dataset.}
  \label{fig:cmc_market}
\end{figure*}

\makeatother
\begin{figure}[t]
  \centering
  \subfigure[CUHK03]{\label{fig:cmc_market4}%
  \includegraphics[width=0.45\textwidth{}]{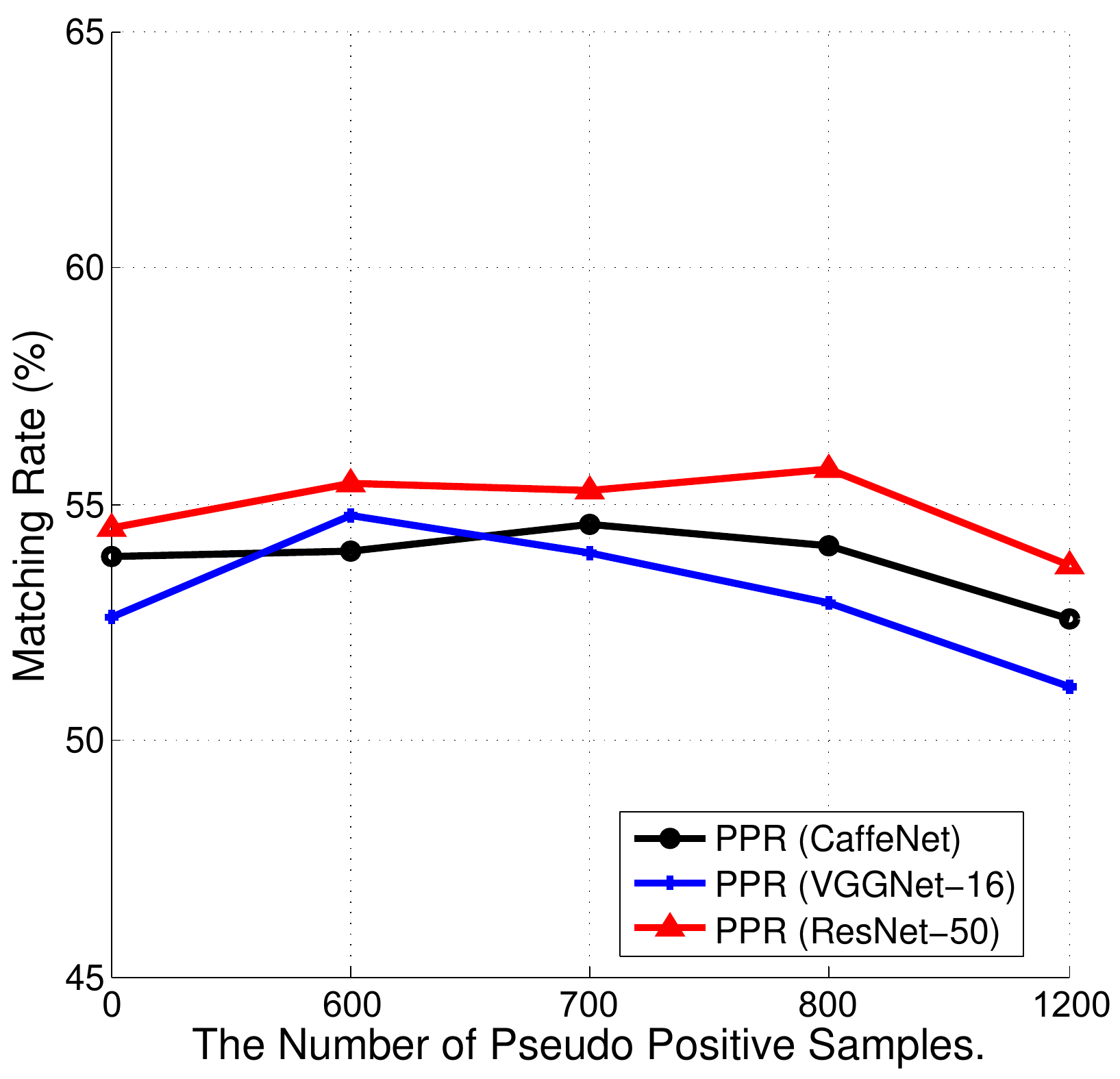}}
  \hspace{0.01in}
  \subfigure[Market-1501]{\label{fig:cmc_market5}%
  \includegraphics[width=0.45\textwidth{}]{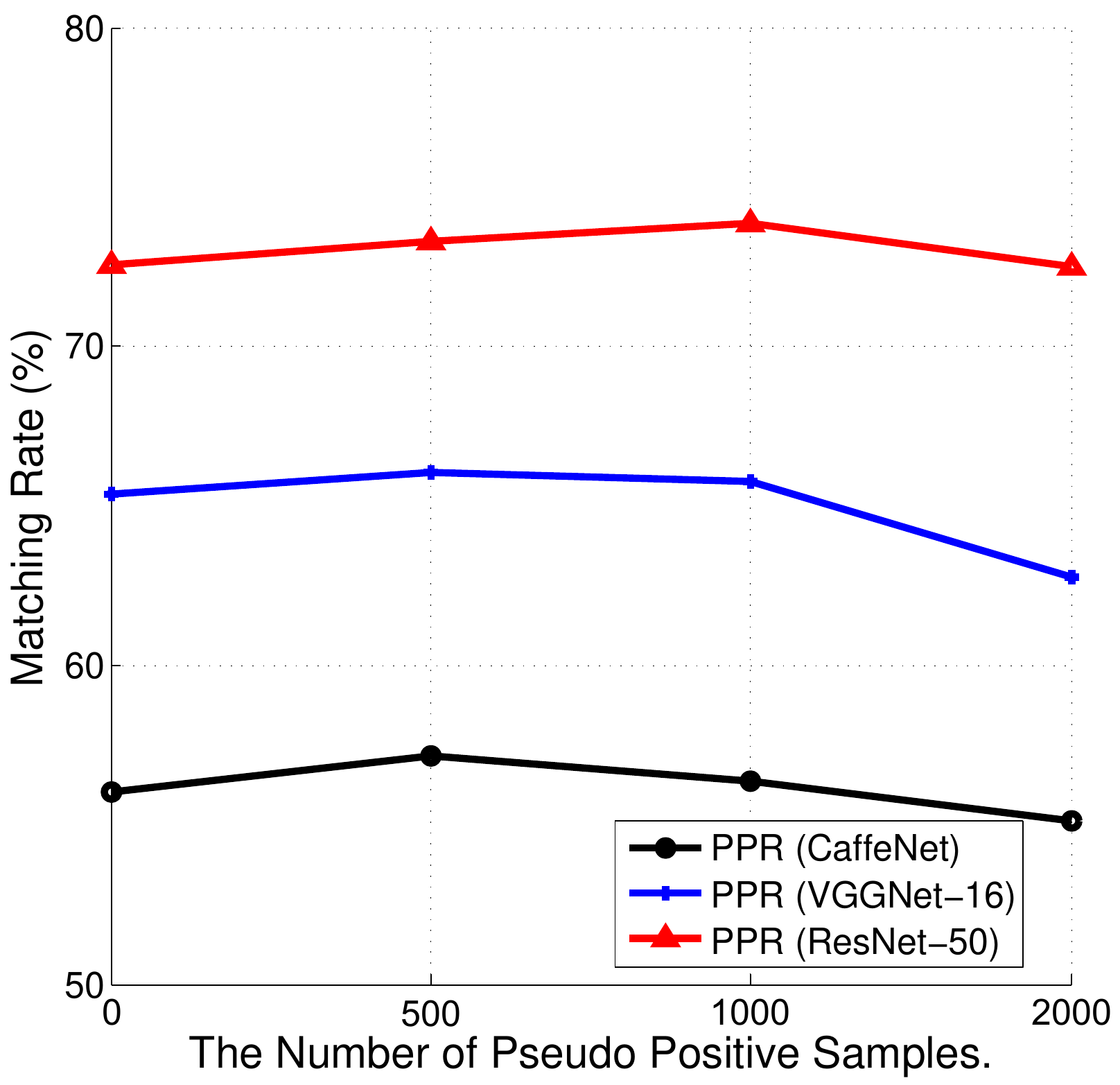}}
  \caption{The rank-1 accuracy curve of the baseline and PPR with different CNN models on CUHK03 and Market-1501 datasets, respectively.}
  \label{fig:r1_cuhkmarket}
\end{figure}

\textbf{The impact of using various numbers of Pseudo Positive samples.} On CUHK03 dataset, the numbers of Pseudo Positive samples are set to 600, 700, 800 and 1,200, respectively. On Market-1501 dataset, the numbers of Pseudo Positive samples are set to 500, 1,000 and 2,000, respectively. We evaluate PPR using various numbers of Pseudo Positive samples with a comparison of the baseline on the two re-ID datasets, and present the results in Table \ref{table:improv_1} and Table \ref{table:improv_2}.

When a relatively small number of Pseudo Positive samples (\emph{i.e.}, 600, 700, and 800 Pseudo Positive samples on CUHK03 dataset, 500 and 1,000 Pseudo Positive samples on Market-1501 dataset.) are added to the original training set, we observe clearly that PPR has a stable improvement compared with the baseline. On Market-1501 dataset, for example, when using ResNet-50 model, PPR exceeds the baseline by +1.33\% (from 72.54\% to 73.87\%) and +1.79\% (from 46.00\% to 47.79\%) in rank-1 accuracy and mAP, respectively. Specifically, the rank-1 accuracy and mAP on Market-1501 dataset arrive at 73.87\% and 47.79\%, respectively. On CUHK03 dataset, similar observation can also be made. It shows that adding a small number of Pseudo Positive samples to the original training set is beneficial to increase the diversity of the training data for a higher discriminative CNN training. Moreover, Fig. \ref{fig:cmc_cuhk} and Fig. \ref{fig:cmc_market} summarize the CMC comparison of the baseline and PPR on CUHK03 and Market-1501 datasets, respectively. From rank-1 to rank-15, the PPR method achieves more excellent matching rate in most cases on the two re-ID datasets. Experimental results suggest that exploiting unlabeled data to generate Pseudo Positive samples is an effective way of reducing the risk of over-fitting to train higher discriminative CNN model for person re-ID task. Fig. \ref{fig:r1_cuhkmarket} summarizes the rank-1 accuracy curve of the baseline and PPR (The horizontal ordinate is the number of Pseudo Positive samples. The baseline is the situation that the number of Pseudo Positive samples is set to 0.) with three types of the CNN models for a more intuitive representation. Generally, the deeper neural network (\emph{i.e.}, VGGNet-16 and ResNet-50) yield higher rank-1 accuracy as shown in Fig. \ref{fig:r1_cuhkmarket} (b) on Market-1501 dataset. On CUHK03 dataset, the rank-1 accuracy of using VGGNet-16 is slightly inferior to the one using CaffeNet model on the situations of 700 and 800 Pseudo Positive samples as shown in Fig. \ref{fig:r1_cuhkmarket} (a). The reason may be that CUHK03 dataset does not contain sufficient training data for VGGNet-16 network.

Remarkably, the re-ID accuracy of PPR does not improve as the number of the Pseudo Positive samples further increase. When we use 1,200 and 2,000 Pseudo Positive samples for evaluation on CUHK03 and Market-1501 datasets, we find that the rank-1 accuracy and mAP have an obvious decrease as shown in Table \ref{table:improv_1} and Table \ref{table:improv_2}. In this case, the re-ID accuracy of PPR is even inferior to the baseline. The reason may be that a large amount of Pseudo Positive samples bring data pollution to the training data. This negative influence on the training set compromises the discriminative ability of the trained CNN model.

\textbf{The comparison with other regularization methods.} In order to further verify the effectiveness of the proposed PPR method for person re-ID, we evaluate another regularization method \textbf{DisturbLabel} \cite{xie2016disturblabel} for comparison on the two re-ID datasets. The DisturbLabel has achieved good performance in the generic image classification task. A proportion of the training labels are replaced as incorrect values during the training process of the CNN model in the DisturbLabel \cite{xie2016disturblabel} method. Yet, the DisturbLabel based regularization strategy dose not bring additional training samples. Beyond that, when we add some noisy samples randomly, this strategy can be regarded as in spirit consistent with DisturbLabel method (denoted as \textbf {DisturbLabel*}). In our experiment, the number of Disturbed samples is set to 700 and 500 on CUHK03 and Market-1501 datasets, respectively. The number of Pseudo Positive samples and Disturbed samples is same for fair comparison. We evaluate the two versions of DisturbLabel method with a comparison of the baseline and PPR on the two re-ID datasets, and present the results in Table \ref{table:improv_3} and Table \ref{table:improv_4}.

Compared with the baseline and the PPR method, we observe clearly that the DisturbLabel and DisturbLabel* have a decrease in the rank-1 accuracy and mAP on the two re-ID datasets. The DisturbLabel is not suitable for person re-ID task. The main reason is that the current data volume of training samples for each identity on re-ID datasets is still far from satisfactory compared with the generic image classification dataset. So randomly disturbing the training labels, even a small proportion, will in effect deteriorate the training process of the CNN. Random labeling does not play an advantageous role during the training process of the CNN model. In our proposed PPR method, the Pseudo Positive samples are generated by visual similarity, which achieves a delicate design to enrich data diversity while avoiding too much data pollution.

\setlength{\tabcolsep}{2.0pt}
\begin{table}[t]
  \centering
  \caption{The rank-1 accuracy (\%) and mAP (\%) comparisons of the baseline, DisturbLabel \cite{xie2016disturblabel} and PPR with different CNN models on CUHK03 dataset. }
  \begin{tabular}{l|cc|cc|cc} \hline
  \multirow{2}{*}{Methods} & \multicolumn{2}{c|}{CaffeNet \cite{krizhevsky2012imagenet}} & \multicolumn{2}{c|}{VGGNet-16 \cite{simonyan2014very}} & \multicolumn{2}{c}{ResNet-50 \cite{he2016deep}}\\\cline{2-7}
  & rank-1& mAP& rank-1& mAP& rank-1& mAP\\ \hline
  Baseline & 53.90 & 60.11& 52.60 & 58.94 & 54.50 & 60.72 \\ \hline
  DisturbLabel \cite{xie2016disturblabel} & 53.45 & 59.89 & 51.90 & 58.37 & 53.95 & 60.13 \\
  DisturbLabel* \cite{xie2016disturblabel} & 53.30 & 59.72 & 52.25 & 58.81 & 53.85 & 60.02 \\ \hline
  \textbf{PPR} & \bf 54.55 & \bf 60.68& \bf 53.95 & \bf 60.48 & \bf 55.30 & \bf 61.42\\ \hline
  \end{tabular}\label{table:improv_3}
\end{table}

\setlength{\tabcolsep}{2.0pt}
\begin{table}[t]
  \centering
  \caption{The rank-1 accuracy (\%) and mAP (\%) comparisons of the baseline, DisturbLabel \cite{xie2016disturblabel} and PPR with different CNN models on Market-1501 dataset. }
  \begin{tabular}{l|cc|cc|cc} \hline
  \multirow{2}{*}{Methods} & \multicolumn{2}{c|}{CaffeNet \cite{krizhevsky2012imagenet}} & \multicolumn{2}{c|}{VGGNet-16 \cite{simonyan2014very}} & \multicolumn{2}{c}{ResNet-50 \cite{he2016deep}}\\\cline{2-7}
  & rank-1& mAP& rank-1& mAP& rank-1& mAP\\ \hline
  Baseline & 56.03 & 32.41& 65.38 & 40.86 & 72.54 & 46.00 \\ \hline
  DisturbLabel \cite{xie2016disturblabel} & 56.29 & 32.28 & 64.13 & 39.50 & 71.56 & 45.70 \\
  DisturbLabel* \cite{xie2016disturblabel} & 56.26 & 31.84& 64.64 & 40.19 & 71.62 & 45.41 \\ \hline
  \textbf{PPR} & \bf 57.16 & \bf 32.60& \bf 66.03 & \bf 41.75 & \bf 73.31 & \bf 47.36\\ \hline
  \end{tabular}\label{table:improv_4}
\end{table}

\setlength{\tabcolsep}{10.0pt}
\begin{table*}[t]
  \centering
  \caption{On CUHK03 and Market-1501 datasets, the rank-1 accuracy (\%) and mAP (\%) comparison of PPR using different quality Pseudo Positive samples. \textbf{C}: Pseudo Positive samples are collected by CaffeNet. \textbf{V}: Pseudo Positive samples are collected by VGGNet-16. \textbf{R}: Pseudo Positive samples are collected by ResNet-50. We use two models for PPR training, \emph{i.e.}, VGGNet-16 and ResNet-50.}
  \begin{tabular}{l|cc|cc|cc|cc} \hline
  \multirow{3}{*}{Methods} & \multicolumn{4}{c|} {CUHK03} & \multicolumn{4}{c}{Market-1501}\\ \cline {2-9} & \multicolumn{2}{c|}{VGGNet-16 \cite{simonyan2014very}} & \multicolumn{2}{c|}{ResNet-50 \cite{he2016deep}}& \multicolumn{2}{c|}{VGGNet-16 \cite{simonyan2014very}} & \multicolumn{2}{c}{ResNet-50 \cite{he2016deep}}\\\cline{2-9}
  & rank-1& mAP& rank-1& mAP& rank-1& mAP& rank-1& mAP\\ \hline
  Baseline & 52.60 & 58.94 & 54.50 & 60.72 & 65.38 & 40.86 & 72.54 & 46.00 \\ \hline
  PPR (C) & 52.75 &59.16 & 54.85 & 60.87 & 65.56 & 40.98& 72.90 & 46.17 \\ 
  PPR (V) & 53.95 & 60.48 & 55.05 & 61.03 & 66.03 & 41.75& 73.07 & 46.56 \\
  PPR (R) & \textbf{54.20} & \textbf{60.59} & \textbf{55.30}& \textbf{61.42} & \textbf{66.48} & \textbf{42.03}& \textbf{73.31} & \textbf{47.36} \\ \hline
  \end{tabular}\label{table:improv_5}
\end{table*}

\textbf{The impact of the retrieved Pseudo Positive samples quality.} As mentioned in Section \ref{sec:improv}, the Pseudo Positive samples are retrieved from the independent database using original supervised training samples as queries. The deep feature of each sample is extracted by the corresponding baseline CNN model (described in Section \ref{sec:base_l}). Due to the differences in network architecture among three types of the baseline CNN models, the quality of the retrieved Pseudo Positive samples is different. Compared with CaffeNet, the VGGNet-16 and ResNet-50 models have higher discriminative ability. It has been indicated by respective results in ILSVRC\footnote{http://www.image-net.org/challenges/LSVRC/} or the references \cite{krizhevsky2012imagenet,simonyan2014very,he2016deep}. Specifically, on ImageNet \cite{deng2009imagenet} validation set, the top-5 classification errors of CaffeNet, VGGNet-16 and ResNet-50 are 16.4\%, 9.33\% and 6.71\%, respectively. The main reason is two-fold. First, VGGNet-16 and ResNet-50 have 16 and 50 layers, respectively, which are deeper than CaffeNet (7 layers). Second, the size of the convolutional kernel is smaller in VGGNet-16 or ResNet-50 models than CaffeNet. In person re-ID, we can also observe in Table \ref{table:base_line} that the re-ID accuracy of VGGNet-16 and ResNet-50 are superior to that CaffeNet on Market-1501 dataset. We view that the quality of Pseudo Positive samples has an influence on the re-ID accuracy in our method. Specifically, we use three models for Pseudo Positive sample collection, \emph{i.e.}, CaffeNet, VGGNet-16 and ResNet-50. We use two models for PPR training, \emph{i.e.}, VGGNet-16 and ResNet-50. In our experiment, the numbers of Pseudo Positive samples are set to 700 and 500 on CUHK03 and Market-1501 datasets, respectively.

\setlength{\tabcolsep}{10.0pt}
\begin{table}
  \centering
  \caption{Comparison with the state-of-the-art person re-ID methods on CUHK03 dataset.}
  \begin{tabular}{l|cc} \hline
  \multirow{2}{*}{Methods} & \multicolumn{2}{c}{CUHK03}\\ \cline{2-3}
  & rank-1& mAP\\ \hline
  BoW+HS \cite{zheng2015scalable} & 24.33 &-\\ 
  BoW+LMNN \cite{weinberger2005distance} & 6.25 &-\\ 
  BoW+ITML \cite{davis2007information} & 5.14 & -\\ 
  BoW+KISSME \cite{koestinger2012large} & 11.70 & -\\ 
  LOMO+XQDA \cite{liao2015person} & 46.25 & -\\ 
  FPNN \cite{li2014deepreid} & 19.89 & -\\ 
  DML \cite{yi2014deep} & 49.84 & -\\ 
  Improved Siamese \cite{ahmed2015improved} & 44.96 & -\\
  SI-CI \cite{wang2016joint} & 52.17 & -\\
  Null Space \cite{zhang2016learning} & 54.70 & -\\ \hline
  PPR (CaffeNet) & 54.55 & 60.68\\ 
  PPR (VGGNet-16) & 54.75 &60.87\\ 
  PPR (ResNet-50) & \textbf{55.75} & \textbf{61.76}\\ \hline
  \end{tabular}\label{table:compare_reid_1}
\end{table}

\setlength{\tabcolsep}{10.0pt}
\begin{table}
  \centering
  \caption{Comparison with the state-of-the-art person re-ID methods on Market-1501 dataset.}
  \begin{tabular}{l|cc} \hline
  \multirow{2}{*}{Methods} & \multicolumn{2}{c}{Market-1501}\\ \cline{2-3}
  & rank-1& mAP\\ \hline
  BoW+HS \cite{zheng2015scalable} & 47.25 &21.88\\ 
  BoW+LMNN \cite{weinberger2005distance} & 34.00 &15.66\\ 
  BoW+ITML \cite{davis2007information} & 38.21 & 17.05\\ 
  BoW+KISSME \cite{koestinger2012large} & 39.61 & 17.73\\ 
  LOMO+XQDA \cite{liao2015person} & 26.07 & 7.75 \\ 
  PersonNet \cite{wu2016personnet} & 37.21 & 18.57\\
  SSDAL \cite{Su2016Deep} & 39.4 & 19.6\\
  TMA \cite{martinel2016temporal} & 47.92 & 22.31\\
  End-to-end CAN \cite{liu2016end} & 48.24  & 24.43\\
  Multiregion Bilinear \cite{ustinova2015multiregion} & 45.58 & 26.11\\ 
  Null Space \cite{zhang2016learning} & 55.43 & 29.87\\
  Siamese LSTM \cite{varior2016siamese} & 61.60 & 35.30\\
  Gated S-CNN \cite{varior2016gated} & 65.88 & 39.55\\ \hline
  PPR (CaffeNet) & 56.35 & 33.11\\ 
  PPR (VGGNet-16) & 66.03 & 41.75\\ 
  PPR (ResNet-50) & \textbf{73.87} & \textbf{47.79}\\ \hline
  \end{tabular}\label{table:compare_reid_2}
\end{table}

The results are listed in Table \ref{table:improv_5}. The Pseudo Positive samples retrieved by higher discriminative CNN model (\emph{i.e.}, VGGNet-16 and ResNet-50) play a more excellent role in PPR. Specifically, taking the results on ResNet-50 model as example, the rank-1 accuracy and mAP increase from 54.85\% to 55.30\%, from 60.87\% to 61.42\% on CUHK03 dataset, respectively. On Market-1501 dataset, the rank-1 accuracy and mAP increase from 72.90\% to 73.31\%, from 46.17\% to 47.36\%, respectively. The reason is that the Pseudo Positive samples generated by higher discriminative CNN model are more similar with original training samples in appearance. These Pseudo Positive samples are more suitable to increase the diversity of training data for reducing the risk of over-fitting during CNN training. The re-ID accuracy of PPR using the Pseudo Positive samples generated by CaffeNet (can be regarded as the relatively weak discriminative Pseudo Positive samples) is also superior to the baseline.

\subsubsection{Comparison with state-of-the-art re-ID methods}
We first compare with the Bag-of-Words (BoW) descriptor \cite{zheng2015scalable}. Here, we only list the best result (\emph{i.e.}, BoW+HS) described in \cite{zheng2015scalable}. In addition, we compare with some existing metric learning methods based on the BoW descriptor \cite{zheng2015scalable} and LOMO feature \cite{liao2015person}. The metric learning methods include LMNN \cite{weinberger2005distance}, ITML \cite{davis2007information}, KISSME \cite{koestinger2012large} and XQDA \cite{liao2015person}. As can be seen in Table \ref{table:compare_reid_1} and Table \ref{table:compare_reid_2}, it is clear that PPR brings decent improvement in both rank-1 accuracy and mAP on the two re-ID datasets. Specifically, on CUHK03 dataset, we achieve rank-1 accuracy = 55.75\%, mAP = 61.76\%. We achieve rank-1 accuracy = 73.87\%, mAP = 47.79\% on Market-1501 dataset.

Then, we compare the proposed method with some state-of-the-art person re-ID methods based on deep learning, including FPNN \cite{li2014deepreid}, DML \cite{yi2014deep}, Improved Siamese \cite{ahmed2015improved}, SI-CI \cite{wang2016joint}, Null Space \cite{zhang2016learning}, PersonNet \cite{wu2016personnet}, Semi-supervised Deep Attribute Learning (SSDAL) \cite{Su2016Deep}, Temporal Model Adaptation (TMA) \cite{martinel2016temporal}, End-to-end Comparative Attention Network (CAN) \cite{liu2016end}, Multiregion Bilinear \cite{ustinova2015multiregion}, Siamese LSTM \cite{varior2016siamese} and Gated S-CNN \cite{varior2016gated}. From the results in Table \ref{table:compare_reid_1} and Table \ref{table:compare_reid_2}, it is clear that PPR significantly outperforms most of deep learning methods in both rank-1 accuracy and mAP by a large margin.

\section{Conclusion}\label{sec_5}
In this paper, we adopt the identification CNN model and propose a \textbf{P}seudo \textbf{P}ositive \textbf{R}egularization (PPR) method to reduce the risk of over-fitting during CNN training for person re-ID. PPR makes full use of the pedestrian identity annotation and enriches the diversity of the original training data by unlabeled data. The problem of the annotation difficulty of training data for each identity can be solved to some extent on the existing re-ID datasets. Experimental results demonstrate that the proposed method provides a stable improvement over the baseline. Compared with the state-of-the-art person re-ID methods, PPR yields a competitive performance on CUHK03 and Market-1501 datasets.

\begin{acknowledgements}
This work was supported in part by the Foundation for Innovative Research Groups of the National Natural Science Foundation of China (NSFC) under Grant 71421001, in part by the National Natural Science Foundation of China (NSFC) under Grant 61502073 and Grant 61429201, in part by the Open Projects Program of National Laboratory of Pattern Recognition under Grant 201407349, and in part to Dr. Qi Tian by ARO grants W911NF-15-1-0290 and Faculty Research Gift Awards by NEC Laboratories of America and Blippar.
\end{acknowledgements}

\bibliographystyle{spmpsci}      
\bibliography{mybibtxt}


\end{document}